\documentclass{article}
\usepackage{spconf,amsmath,graphicx}
\usepackage{url}
\usepackage{booktabs,multirow}
\usepackage{makecell}
\usepackage{hyperref}

\newcommand\blfootnote[1]{%
  \begingroup
  \renewcommand\thefootnote{}\footnote{#1}%
  \addtocounter{footnote}{-1}%
  \endgroup
}

\title{Semi-Supervised Training for Improving Data Efficiency in End-to-End Speech Synthesis}

\name{Yu-An Chung\thanks{$^{*}$Work done while at Google.}$^{*1}$\quad Yuxuan Wang$^{2}$\quad Wei-Ning Hsu$^{*1}$ \quad Yu Zhang$^{2}$\quad RJ Skerry-Ryan$^{2}$}
\address{$^{1}$Massachusetts Institute of Technology\qquad $^{2}$Google Inc.}

\begin{document}

\maketitle

\begin{abstract}
Although end-to-end text-to-speech~(TTS) models such as Tacotron have shown excellent results, they typically require a sizable set of high-quality $<$text, audio$>$ pairs for training, which are expensive to collect.
In this paper, we propose a semi-supervised training framework to improve the data efficiency of Tacotron.
The idea is to allow Tacotron to utilize textual and acoustic knowledge contained in large, publicly-available text and speech corpora.
Importantly, these external data are unpaired and potentially noisy.
Specifically, first we embed each word in the input text into word vectors and condition the Tacotron encoder on them.
We then use an unpaired speech corpus to pre-train the Tacotron decoder in the acoustic domain.
Finally, we fine-tune the model using available paired data.
We demonstrate that the proposed framework enables Tacotron to generate intelligible speech using less than half an hour of paired training data.

\end{abstract}


\section{Introduction}
\label{sec:intro}
Recent advances in end-to-end text-to-speech~(TTS) have shown great promise.
We are now able to produce natural prosody with high audio fidelity using a much simplified voice building pipeline~\cite{wang2017tacotron,shen2018natural,ping2018clarinet}.
However, such models typically require a sizable dataset consisting of high-quality $<$text, audio$>$ training pairs, which are expensive and time-consuming to collect.
Requiring large amounts of data also hinders their applicability in low-resource settings.
\blfootnote{Sound demos can be found at \url{https://google.github.io/tacotron/publications/semisupervised}.}

This work aims to improve the data efficiency for end-to-end TTS training by leveraging large-scale, publicly available, and \textit{unpaired} text and speech data.
Unpaired data are plentiful and relatively easy to collect.
Specifically, we propose a simple yet effective semi-supervised framework for training Tacotron~\cite{wang2017tacotron}, a recently proposed end-to-end TTS model.
We propose to transfer the textual and acoustic representations learned from unpaired data to Tacotron in an unsupervised manner.
This is then followed by a fine-tuning step using only a small amount of paired data to learn the alignment between the two representation domains.

In this preliminary study, we first identify the data requirement of a baseline Tacotron, i.e., the least amount of training data needed for a baseline Tacotron to produce intelligible speech.
We then show that a Tacotron enhanced with the proposed framework is able to produce intelligible speech using less amount of data.
Finally, we study different configurations for incorporating the framework.
For evaluation, we perform both objective and subjective tests.

There exists previous work studying the application of unsupervised and weakly supervised learning for TTS~\cite{watts2013unsupervised,lu2013combining,watts2015sentence, wang2015word}.
Related to our work, for example, \cite{wang2015word} uses pre-trained word vectors in a LSTM-based acoustic model in parametric TTS~\cite{wang2015word}.
These studies consider learning methods within the traditional TTS paradigm, however.
This work, by contrast, examines them within end-to-end TTS, and specifically targets the data efficiency problem.


\section{Proposed Approach}
We use a baseline Tacotron architecture specified in~\cite{wang2018style}, where we use a GMM attention~\cite{graves2013generating}, LSTM-based decoder with zoneout regularization~\cite{krueger2017zoneout} and phoneme inputs derived from normalized text.
We use Griffin-Lim~\cite{griffin1984signal} as the inversion algorithm to convert the predicted spectrograms to waveforms, as our main focus is to enable Tacotron training on small data instead of producing high-fidelity audio.
Using Griffin-Lim allows much faster experiment cycles.

The two main building blocks of Tacotron are the encoder and the attention-based decoder.
At a high level, the encoder takes a source text as input and produces sequential representations of it; the decoder is then conditioned on the text representations to generate corresponding acoustic representations~(spectrogram frames), which are then converted to waveforms.
In the baseline Tacotron, the model is trained from scratch where all network weights are randomly initialized, and both the text and acoustic representations are learned from the given~(parallel) training data.
Below, we introduce our approach to inject external textual and acoustic knowledge to bootstrap the encoder and decoder, respectively.

\subsection{Conditioning the encoder on pre-trained word vectors}
\begin{figure}[t]
  \begin{minipage}[b]{1.0\linewidth}
    \centering
    \centerline{\includegraphics[width=8.5cm]{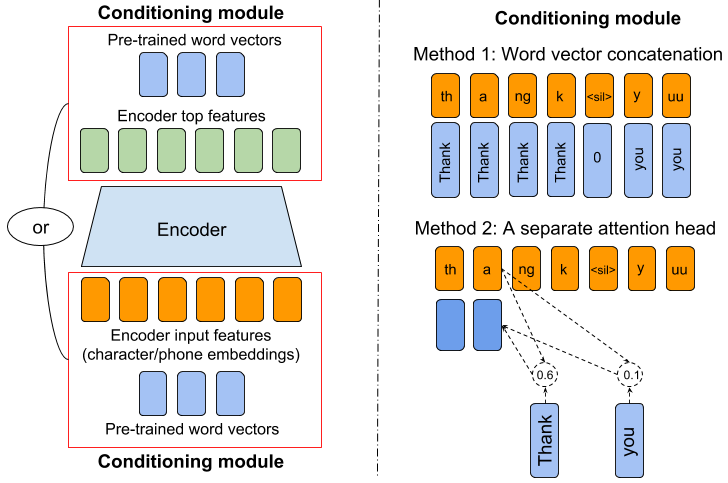}}
  \end{minipage}
  \caption{
    Illustration of conditioning encoder on pre-trained word vectors.
    The left side shows the locations of encoder input and encoder top, where the word vectors can be incorporated via a conditioning module.
    The right side illustrates the conditioning module, where we show two methods of conditioning.
    `$<$sil$>$' denotes silence~(e.g. space).
    In method~2, the dash lines correspond to the attention mechanism.}
  \label{fig:encoder-example}
\end{figure}
The goal of the encoder is to extract robust sequential representations of text.
However, for a baseline Tacotron, the only training signal comes from the text data in the $<$text, audio$>$ pairs, and the extracted representations are usually not rich enough when there's only a small amount of text.

We propose to exploit the textual knowledge contained in large text corpora, which typically contain millions to billions of words.
From these large text corpora, one can train real-valued word vectors that contain the meanings of the words~\cite{mikolov2013distributed,pennington2014glove} or language models that model grammatical and semantic context~\cite{bengio2003neural}.
These word vectors can be added as auxiliary inputs to a TTS model to convey additional textual knowledge not learnable from the original text data.

To expose this additional knowledge to the encoder, we first embed each word in the input text into a word vector, and add the word vector sequence on one of two locations~(illustrated in the left side of Figure~\ref{fig:encoder-example}): ``encoder input'' representing the phoneme embedding sequence, or ``encoder top'' representing the final encoder output sequence.
While both conditioning locations allow the encoder to access the pre-trained word vectors, the choice of the conditioning location is an important design choice, which we study in experiments.
For convenience, we call both encoder input and encoder top features \textit{conditioning location features}.
Due to the fact that the word vectors and conditioning location features might have different time resolutions~(different sequence lengths), below we propose two ways of combining them~(illustrated in the right side of Figure~\ref{fig:encoder-example}).

\subsubsection{Word vectors concatenation}
The first conditioning approach concatenates the word vector at the first phoneme of the corresponding word and replicates the word vector across all phonemes in the word.
Take input text ``Thank you'' as an example.
The phoneme inputs are `th', `a', `ng', `k', `$<$sil$>$', `y', `uu'~(same rule applies to character inputs), where `$<$sil$>$' denotes silence~(e.g. space).
The word vector of ``Thank'' is appended to the phoneme embeddings of phonemes `th', `a', `ng', and `k'.
This approach can be thought of as an hard attention mechanism where the alignment is pre-determined by the position mapping between words and their phonemes in the input text.

\subsubsection{A conditioning attention head}
If we consider the phrase ``Thank you'', it's possible that the semantics of ``Thank'' can help the encoder to generate more robust representation for ``you''.
However, the first approach never exposes the word vector of ``Thank'' to the encoder when it's processing the phoneme embeddings of ``you''.
Our second approach attempts to resolve this by applying a separate attention head between word vectors and conditioning location features.
It takes each conditioning location feature as the attention query to generate the corresponding context vector, which is a weighted sum of the word vectors.
The context vector and the conditioning location feature are then concatenated together for further processing.
This enables each conditioning location feature to extract and gather information it needs from all word vectors.
In this work, we use a simple tanh based additive attention~\cite{bahdanau2015neural}.

Since the encoder weights are still trained from scratch with random initialization, we refer to this approach as \textit{encoder conditioning} for the rest of the paper.

\subsection{Decoder pre-training}
In a baseline Tacotron system, the decoder needs to simultaneously learn acoustic representations and their alignments with the text representations extracted by the encoder.
To reduce the workload of the decoder, we propose using an independent speech data source to pre-train the decoder, such that it is initialized by a pre-learned acoustic representation.
During pre-training, the decoder acts as a next-step frame predictor with teacher forcing.
Since the only objective is to predict an acoustic frame from the previous one, this step does not require text transcripts.
In this stage, we simply keep the encoder weights frozen and replace the attention context vectors by zero vectors.
This forces the decoder to learn an autoregressive model of acoustics at the frame level.

After the decoder is pre-trained, we fine-tune the entire model~(including both encoder and decoder) using paired data.
By pre-training, the decoder no longer needs to learn the acoustic representations from scratch and can thus focus more on learning the alignment between text and acoustic representations.

A potential source of error of our simple approach is that there is a model mismatch between decoder pre-training and model fine-tuning: during pre-training, the decoder is only conditioned on the previous frame; while during fine-tuning, it is additionally conditioned on the text representations from the encoder.
Despite such a mismatch, we found decoder pre-training still helpful.
In addition, we found that the pre-trained Tacotron converges much faster than the baseline.

\section{Experiments}
We conduct experiments to demonstrate the effectiveness of our framework.
We use an internal single-speaker US English dataset for training~(fine-tuning).

\subsection{Data requirements of the baseline Tacotron}
To improve Tacotron's data efficiency, first we need to understand its limit.
We'd like to answer the following question: what is the maximum amount of data~$N$ that could almost never successfully train a baseline Tacotron to produce intelligible speech?
To find out~$N$, we gradually decrease the amount of data used for training a baseline Tacotron from about 40 hours to about 12 minutes and listen to the synthesized speech on unseen phrases.
As can be heard on our \href{https://google.github.io/tacotron/publications/semisupervised}{demo page}, we estimated that using between~10 and~40 hours of data produces almost equally good synthesis, and using between~3 and~10 hours of data causes minor degradation but still sounds very good.
However, when there are only about~24 minutes of data, the model fails to produce intelligible speech.
When there are only~12 minutes of data, the model outputs gibberish that is impossible to understand.
It's important to note that the transcripts in the~12 minutes data already cover all phonemes, therefore the failure is not simply due to phoneme coverage.

Therefore, in the next section, we focus on demonstrating the effectiveness of our semi-supervised framework using only~24 minutes of paired data.

\subsection{Results on small data}
Our encoder conditioning and decoder pre-training approaches can be applied to Tacotron independently or jointly.
We denote the model that only incorporates encoder conditioning as \textit{T-Enc}, model that only incorporates decoder pre-training as \textit{T-Dec}, model that incorporates both as \textit{T-Enc-Dec}, and the baseline Tacotron as \textit{T-Base}.

We measure the synthesis quality using both objective and subjective tests.
For the objective metric, we use mel cepstral distortions~(MCD)~\cite{kubichek1993mel}, which measures the distance between synthesis and ground truth in the mel cepstrum space---the smaller the better.
We use an evaluation set containing about~30 minutes~(631 sentences) of unseen data.
We found that our MCD results correlate well with our subjective perception.
For subjective measurements, we ran a series of side-by-side preference tests using~1000 unseen phrases of different lengths.

For encoder conditioning, we used a neural network language model~(NNLM)~\cite{bengio2003neural} trained on English Google News 200B corpus from TensorFlow Hub as the word embedding module.
The module maps each word to a 128-dimensional vector.
We also tried word2vec~(W2V)~\cite{mikolov2013efficient} trained on the same corpus as the word embedding module.

For decoder pre-training, we used VCTK~\cite{veaux2017cstr}, a publicly available corpus containing~44 hours of speech from~109 speakers, the majority of which have British accents.
Note that there is an accent mismatch between the decoder pre-training~(multiple speakers with British accents) and fine-tuning~(single speaker with US accent) datasets.
As mentioned above, we only use the speech signals in VCTK but not their transcripts.

\subsubsection{MCD objective tests}
\begin{table}[t]
  \centering
  \caption{
    MCD between ground-truth audio and synthesis from~7 Tacotron variants~(lower is better).
    For T-Enc, we include both the results of using NNLM~(1st row) and W2V~(2nd row) as the word embedding module; concatenation/attention and input/top denote the conditioning method and location, respectively.
    The best result is marked in bold.}
  \label{tab:mcd-results}
  \resizebox{\columnwidth}{!}{
  \begin{tabular}{ccccccc}
    \toprule
    \multirow{3}{*}{T-Base}  &  \multicolumn{4}{c}{T-Enc}  &  \multirow{3}{*}{T-Dec}  &  \multirow{3}{*}{T-Enc-Dec} \\
    \cmidrule(lr){2-5}
    &  \multicolumn{2}{c}{concatenation}  &  \multicolumn{2}{c}{attention} & &\\
    \cmidrule(lr){2-3}  \cmidrule(lr){4-5}
    &  input  &  top  &  input  &  top & &\\
    \midrule
    18.06  &  \makecell{12.89 \\ 13.72}  &  \makecell{12.46 \\ 13.14}  &  \makecell{13.03 \\ 13.86}  &  \makecell{12.71 \\ 13.51}  &  \textbf{12.09}  &  12.27 \\
    \bottomrule
  \end{tabular}
  }
\end{table}
The MCD results are shown in Table~\ref{tab:mcd-results}.
We first compare the four configurations of encoder conditioning.
Here we include the results of both NNLM and W2V word embeddings.
We can see that models using NNLM always outperform their W2V counterpart.
We speculate that this is because W2V only conveys word meanings but not the contextual or structural information, which is modeled in NNLM.

In terms of conditioning locations, we see that conditioning at encoder top always outperforms conditioning at encoder input.
While feeding word vectors to the early parts of the network seems intuitive~(as in encoder input conditioning), we believe it is not the best choice in the low-resource setting.
If the encoder weights learned from small data are noisy, for example, they may ``distort'' well-trained word vectors.
Therefore, conditioning word vectors at a higher layer~(e.g. encoder top) may lead to better generalization.

In terms of conditioning method, we find that the simple concatenation method always outperforms using a separate attention head.
We also attribute this to the limited training data: although a separate attention head offers more flexibility for learning the alignment, it also introduces more trainable parameters.
In summary, the best configuration for encoder conditioning is to directly concatenate word vectors obtained from a pre-trained NNLM at the encoder top.
We used this configuration for T-Enc for the rest of the experiments.

From Table~\ref{tab:mcd-results} we can see that T-Enc, T-Dec, and T-Enc-Dec all achieve much lower MCD than T-Base.
Among them, T-Dec achieves the best result.
However, T-Enc, T-Dec, and T-Enc-Dec achieve similar MCD results~(12.46, 12.09, 12.27, respectively).
As shown in side-by-side comparisons below, the raters did not strongly prefer one over the other two, either.

\subsubsection{Side-by-side subjective tests}
\begin{table}[htbp]
  \centering
  \caption{
    Results of SxS subjective tests based on a 7-point rating scale.
    We report both rater preferences~(in percentage) and $p$-values for each comparison.}
  \label{tab:sxs-results}
  \resizebox{\columnwidth}{!}{
  \begin{tabular}{c|cccc}
    \toprule
    \multirow{2}{*}{\makecell{Competing pair}}  &  \multicolumn{3}{c}{Preference~(\%)}  & \multirow{2}{*}{$p$-value} \\
    \cmidrule(lr){2-4}
       &  Former  &  Latter  &  Neutral  &  \\
    \midrule
    \makecell{T-Base vs. T-Enc}  &   3.3  &  65.1  &  31.6  &  1.07e-84 \\
    \makecell{T-Base vs. T-Dec}  &   3.2  &  61.8  &  35.0  &  3.47e-83 \\
    \makecell{T-Enc vs. T-Dec}   &  16.1  &  18.2  &  65.7  &  0.256    \\
    \makecell{T-Enc-Dec vs. T-Dec}  &  17.0  &  17.9  &  65.1  &  0.630    \\
    \bottomrule
  \end{tabular}
  }
\end{table}
Table~\ref{tab:sxs-results} shows the results of the four side-by-side~(SxS) preference tests, comparing T-Base against T-Enc, T-Base against T-Dec, T-Enc against T-Dec, and T-Enc-Dec against T-Dec.
As we can see from the table, both T-Enc and T-Dec significantly outperform T-Base: in both tests, raters strongly preferred them over the baseline by more than 60\%.
Interestingly, the raters considered T-Dec, T-Enc, and T-Enc-Dec similarly preferable.
The results of SxS tests are consistent to those of MCD objective tests, and both demonstrate the effectiveness of our semi-supervised framework.

\subsection{Results on other amounts of data}
\begin{figure}[t]
\begin{minipage}[b]{1.0\linewidth}
  \centering
  \centerline{\includegraphics[width=6.7cm]{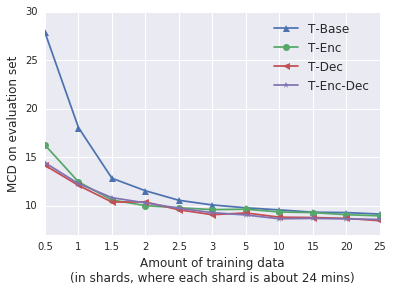}}
\end{minipage}
\caption{MCD results by increasing the amount of paired data.}
\label{fig:varying-data}
\end{figure}
We also compare T-Base, T-Enc, T-Dec, and T-Enc-Dec trained on other amounts of data.
In Figure~\ref{fig:varying-data}, each curve corresponds to a Tacotron variant, showing the relationship between the amount of paired data used for training that Tacotron variant and the MCD between ground-truth audio and synthesis from it.
We can see that the largest gap between T-Base and the three semi-supervised systems occurs when using only 12 minutes~(0.5 shards) of paired data.
The gap keeps decreasing when the amount of data increases.
This phenomenon is somewhat expected, because with more paired data, Tacotron relies less on external knowledge for learning representations and alignments.
However, semi-supervised Tacotron consistently achieves lower MCD than the baseline, which may indicate benefits beyond better data efficiency~(e.g. improved prosody).

\section{Conclusions and Discussions}
We have proposed a semi-supervised training framework for improving data efficiency in end-to-end TTS.
Our framework leverages large-scale, publicly available, and unpaired text and speech data to provide additional textual and acoustic knowledge to the Tacotron encoder and decoder, respectively.
We have shown that our framework makes end-to-end TTS feasible in small-data regime.
Specifically, a semi-supervised trained Tacotron can produce intelligible speech using just 24 minutes of paired training data.
This promising result also provides some guiding principles for future data collection efforts for both single and multi-speaker TTS.
While we used Tacotron as the TTS model in this study, we believe the framework is generally applicable to other end-to-end TTS models.

This is only a preliminary work, and there is still much to be investigated.
For example, we've been using phoneme inputs in this work and we'd like to understand the performance tradeoffs on grapheme inputs.
For leveraging textual knowledge, instead of simply conditioning with word vectors, a likely more effective method is to initialize the entire encoder with a pre-trained bidirectional NNLM~\cite{peters2018deep}.
For decoder pre-training, the model mismatch during pre-training and fine-tuning can be further studied.
An analysis on what kind of information are extracted from external data and how they are actually used by Tacotron is also an important future work.
Lastly, since the main focus of this work is to make end-to-end TTS feasible in small-data regime instead of producing high-fidelity audio, we only used Griffin-Lim as the waveform synthesizer.
To produce high-fidelity speech with very little paired data, we still need to address the problem of adapting neural vocoders in the semi-supervised setting.

\section{Acknowledgements}
The authors thank Daisy Stanton, Eric Battenberg, Soroosh Mariooryad, Yinfei Yang, and the Machine Hearing and Google Brain teams for their helpful feedback and discussions.

\vfill\pagebreak

\bibliographystyle{IEEEbib}
\bibliography{strings,refs}

\end{document}